\documentclass[sigconf]{acmart}
\AtBeginDocument{%
  \providecommand\BibTeX{{%
    \normalfont B\kern-0.5em{\scshape i\kern-0.25em b}\kern-0.8em\TeX}}}

\setcopyright{acmcopyright}
\copyrightyear{2022}
\acmYear{2022}
\acmDOI{10.1145/3557918.3565869}

\acmConference[SIGSPATIAL ’22]{ACM SIGSPATIAL International Conference on Advances in Geographic Information Systems}{November 1-4, 2022}{Seattle, WA, USA}

\acmPrice{15.00}
\acmISBN{978-1-4503-9529-8/22/11}

\usepackage{subcaption}

\begin{document}

\title{Real-time GeoAI for High-resolution Mapping and Segmentation of Arctic Permafrost Features}
\subtitle{The case of ice-wedge polygons}

\author{Wenwen Li}
\authornote{Corresponding Author}
\affiliation{%
  \institution{School of Geographical Sciences and Urban Planning, Arizona State University}
  \city{Tempe}
  \state{AZ}
  \country{USA}
  \postcode{85287-5302}
}
\email{wenwen@asu.edu}

\author{Chia-Yu Hsu}
%\orcid{0000-0002-8923-1213}
\affiliation{%
  \institution{School of Geographical Sciences and Urban Planning, Arizona State University}
  \city{Tempe}
  \state{AZ}
  \country{USA}}
\email{chsu53@asu.edu}

\author{Sizhe Wang}
\affiliation{%
  \institution{School of Computing and Augmented Intelligence, Arizona State University}
  \city{Tempe}
  \state{AZ}
  \country{USA}}
\email{wsizhe@asu.edu}

\author{Chandi Witharana}
\affiliation{%
  \institution{Department of Natural Resources and the Environment, University of Connecticut}
  \city{Storrs}
  \state{CT}
  \country{USA}}
\email{chandi.witharana@uconn.edu}

\author{Anna Liljedahl}
\affiliation{%
  \institution{Woodwell Climate Research Center}
  \city{Falmouth}
  \state{MA}
  \country{USA}}
\email{aliljedahl@woodwellclimate.org}

\renewcommand{\shortauthors}{Li, et al.}

\begin{abstract}
This paper introduces a real-time GeoAI workflow for large-scale image analysis and the segmentation of Arctic permafrost features at a fine-granularity. Very high-resolution (0.5m) commercial imagery is used in this analysis. To achieve real-time prediction, our workflow employs a lightweight, deep learning-based instance segmentation model, SparseInst, which introduces and uses Instance Activation Maps to accurately locate the position of objects within the image scene. Experimental results show that the model can achieve better accuracy of prediction at a much faster inference speed than the popular Mask-RCNN model. 
\end{abstract}




\maketitle

\section{Introduction}
Polar regions are one of Earth’s remaining frontiers that play a vital role in global climate, ecosystems, and economy. Global warming over the past century is driving dramatic change in the Arctic ecosystem, endangering its natural environment, infrastructure, and life of the indigenous population. Permafrost, ground that remains below 0\textdegree C for at least two consecutive summers, is at the center of this change. Covering nearly $\frac{1}{4}$ of the land in the northern hemisphere, thawing permafrost is causing significant local and regional impacts on the Arctic community. As the ice-rich frozen ground thaws, land subsides causing severe damage to buildings, roads, pipelines, and industrial infrastructure \cite{4_hjort2018degrading}. Permafrost degradation also increases rates of coastal erosion, wildfires, and flooding, which may further accelerate the thawing process and make the Arctic ecosystem even more vulnerable to climate change \cite{8_grosse2016changing}. At a global scale, the thawing of Arctic permafrost will result in the release of an immense amount of carbon dioxide and methane, exaggerating the greenhouse effect and global warming through complex feedback mechanisms \cite{schuur2015climate}.

To improve our understanding of permafrost dynamics and its linkages to other Arctic ecosystem components in the midst of rapid Arctic change, it is critically important to have spatial data readily available that provide fine-granularity mapping of permafrost features, their extent, distribution, and longitudinal changes. Achieving this goal requires new approaches that can perform automated mining from Arctic big data. It is exciting that the Arctic community has started to embrace GeoAI \cite{li2020geoai,li2022geoai} and big data to support Arctic research, from predicting Arctic sea ice concentration \cite{13_andersson2020deep}, to finding marine mammals on ice \cite{14_NOAA}, creating Arctic land cover maps \cite{15_song2019patch}, and automated mapping of permafrost features \cite{17_bhuiyan2020use}. Pioneering research in performing automated characterization of Arctic permafrost features has also been reported in the literature. An GeoAI-based Mapping Application for Permafrost Land Environment (MAPLE) is being developed to integrate Big Imagery, GeoAI, and High-Performance Computing (HPC) to achieve classification of permafrost features, in particular, ice-wedge polygons (IWP) \cite{udawalpola2022optimal}. The delineation of IWPs is achieved using a popular instance segmentation model, Mask R-CNN \cite{maskrcnnhe2017mask}. \citet{lingcao2022accuracy} applied a semantic segmentation model U-Net for mapping retrogressive thaw slumps, another important feature type of Arctic permafrost for understanding permafrost thaw and Arctic warming. 
  
While these deep learning models, such as Mask R-CNN, result in satisfying performance in terms of prediction accuracy, they can hardly achieve real-time processing because the algorithms often require placement of a large number of candidate bounding boxes and complex post-processing to remove redundant information. To reduce computational cost and perform efficient permafrost mapping at the pan-Arctic scale (which covers over 5 million $\text{km}^2$ of tundra region), it is necessary to develop and apply new models that can achieve high-accuracy and real-time prediction. This paper aims to achieve this goal by integrating a novel real-time instance segmentation model, SparseInst \cite{cheng2022sparse}, in our automated permafrost feature mapping pipeline. The next section describes the methodological workflow in detail. 

\begin{figure}[h]
  \centering
  \includegraphics[width=0.95\linewidth]{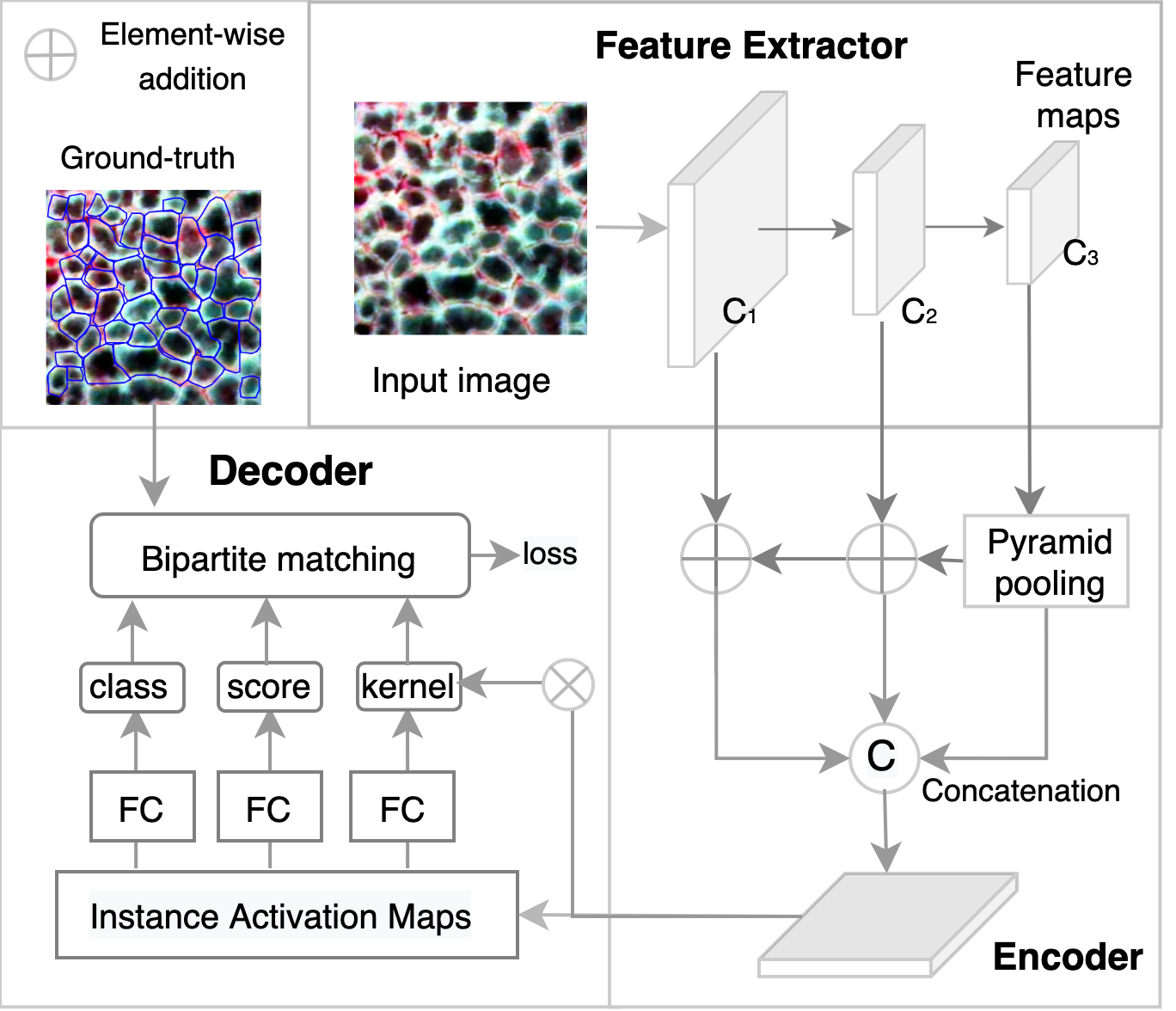}
  \caption{Real-time GeoAI workflow for Arctic permafrost segmentation and mapping. FC: Fully Connected layer}
  \Description{}
  \label{fig:arch}
\end{figure}

\section{Method}
Figure \ref{fig:arch} demonstrates the workflow of real-time GeoAI for Arctic permafrost mapping. We adopt a novel instance segmentation model SparseInst into the workflow, which contains three major components: a feature extractor, an instance context encoder, and an Instance Activation Map (IAM)-based decoder. The feature extractor is responsible for extracting multi-scale features from the input. The encoder will process the extracted features and fuse them into single-level features with multi-scale representations. The encoded features are then processed by the decoder to generate IAMs for instance classification and segmentation. Each component is designed under the consideration of lightweight architecture and low computational complexity to achieve fast inference speed. 

\subsection{Feature Extractor} \label{sec:feature_extractor}
The feature extractor adopted in this work is ResNet-50 \citep{he2016deep}. Among various deep neural network (DNN) architectures, ResNet-50 enjoys a good trade-off between accuracy and model complexity so to support real-time applications \citep{bianco2018benchmark}. ResNet extracts representative features for objects of different types using a deep residual network. After a series of convolutional operations, multi-scale feature maps can be generated, among which high-resolution maps are better at small-object segmentation and low-resolution feature maps can better support segmentation of large objects. To accurately segment objects of varying sizes,  hierarchical feature maps at multiple scales and resolutions are passed to the encoder (see Figure \ref{fig:arch}).  

\subsection{Instance Context Encoder}
The main purpose of the encoder is to generate a single feature map containing multi-scale representations. Conventional approaches use multi-scale features with multi-level predictions \citep{li2019scaleaware} for segmenting objects at different scales \citep{NEURIPS2020_cd3afef9}. However, this will increase overall processing time of the model, making it less efficient and less favorable for real-time applications. Recent real-time instance segmentation models \citep{perreault2021centerpoly,bolya2019yolacta} fuse multi-scale information into a single feature map to reduce both prediction and post-processing time. SparseInst utilizes a similar idea and it fuses three feature maps obtained from different convolution stages. The fusion first follows the feature pyramid network (FPN) \citep{lin2017feature} to use a top-down pathway for building semantic-rich features. To further enhance the scale information, the last feature map ($\text{C}_3$) also undergoes a pyramid pooling operation \citep{zhao2017pyramida} to increase the global contextual information without increasing the size of the feature maps. Next, all feature maps are upsampled to the same resolution and concatenated together to generate feature maps at a single resolution but with multi-scale representations. The output is then sent to the decoder for classification and segmentation. 

\subsection{IAM-based Decoder} 
The function of the decoder is to take the fused feature map from the encoder as input to generate $N$ predictions. Each prediction contains a triple <object class, objectness score, kernel>. The objectness score refers to the probability of an object belonging to a certain class and the kernel is a low-dimensional representation of location information for that object. This instance-level prediction is achieved through the generation of Instance Activation Maps (IAMs) which are capable of highlighting important image areas. Different from conventional approaches which use dense anchors to detect and segment objects, SparseInst trains the decoder to create IAMs, which have a one-to-one mapping with the objects to segment. This design helps the decoder to achieve real-time performance as it avoids the time-consuming post-processing of some models, such as Mask R-CNN, which need to select from thousands of anchors to predict the most accurate mask and to perform matching between predicted masks and the ground-truth. Once the predictions are generated, they are sent to perform bipartite matching to associate each ground-truth object with its most similar prediction, then the difference between the prediction and the ground-truth is encoded into the loss function. As the model is being trained, it learns to generate more accurate IAMs and thus more accurate predictions, lowering the loss until the model fully converges. 

\section{Experiments and Results}

\subsection{Data}
To assess the performance of the models, we created an AI-ready dataset containing 867 image tiles and a total of 34,931 ice-wedge polygons (IWPs). The dataset covers dominant tundra vegetation types in the polygonal landscapes, including sedge, tussock, and barren tundra. Very high resolution (0.5 m) remote sensing imagery acquired by Maxar sensors is used for annotation and model training. The average image size is $\sim 226\times 226$ with the largest image size $507\times 507$. Each image has a label indicating the image size and coordinates of the IWPs.  

The labeled images are divided into three sets: training (70\%), validation (15\%), and testing (15\%). The maximum number of IWPs per image is 447. This statistic is critical in determining the maximum number of detections per image, as it is an important hyperparameter to set in the segmentation model. It also affects both accuracy and speed and provides a trade-off between them (Section \ref{exp:pvs}). 

\begin{comment}

\begin{table}
  \caption{Dataset Statistics.}
  \label{tab:data_sta}
  \begin{minipage}{\columnwidth}
  \begin{tabular}{lccc}
    \toprule
    Set & Number & Total & Number of IWPs/image \\ 
    \newline & of images & IWPs & (min/max) \\
    \midrule
    Training & 606 & 23770$^*$ & 0/447 \\
    Validation & 129 & 6045 & 0/313 \\
    Testing & 132 & 5116 & 5/285 \\
  \bottomrule
\end{tabular}
\footnotesize{\text{*One polygon is incorrectly labeled.}}
\end{minipage}
\end{table}
\end{comment}

\begin{table}
  \caption{Comparisons with Mask R-CNN \citep{maskrcnnhe2017mask} for mask AP and speed on IWP dataset. Inference speeds of all models are tested with single NVIDIA A5000 GPU. }
  \label{tab:comparison}
  \begin{tabular}{lccccc}
    \toprule
    Model & FPS & AP$_{50}$ & AP$_S$ & AP$_M$ & AP$_L$  \\
    \midrule
    Mask R-CNN & 27.01 & 52.86 & 33.28 & 60.03 & 64.39 \\
    SparseInst & 45.61 & 53.97 & 31.70 & 60.78 & 68.10 \\

  \bottomrule
\end{tabular}
\end{table}

\subsection{Model Training and Results}
In this work, we compare SparseInst with one of the most popular instance segmentation models, Mask R-CNN \citep{maskrcnnhe2017mask}. Both models are built upon Detectron2 \citep{wu2019detectron2}, a module of the PyTorch deep learning framework which provides state-of-the-art segmentation algorithms. The training is conducted on four NVIDIA A5000 GPUs. The batch size is 16 and the maximum number of iterations is 20,000. The maximum number of detections per image $N$ is set to 500. Table \ref{tab:comparison} shows the performance comparison between Mask R-CNN (default setting) and SparseInst. The evaluation metric for model inference speed is frame per second (FPS) and for accuracy, average precision (AP) \citep{Zhang2009} is used. As the results show, SparseInst demonstrates better performance in terms of both speed and accuracy than Mask R-CNN. We also separate IWPs into three groups by their areas: small (area < 200 pixels), medium (area in between 200 and 450 pixels), and large (area > 450 pixels). Table \ref{tab:comparison} also shows the average precision (AP) in each group. SparseInst performs slightly worse than Mask R-CNN on small IWPs segmentation, but it works better at segmenting medium- to large-size IWPs. Overall, SparseInst yields better detection accuracy than Mask R-CNN. Speed-wise, the model runs nearly twice as fast as Mask R-CNN, achieving real-time performance (model's inference speed at 30 FPS or above).

\subsection{Precision vs. Speed}\label{exp:pvs}
\begin{figure}[h]
  \centering
  \includegraphics[width=0.9\linewidth]{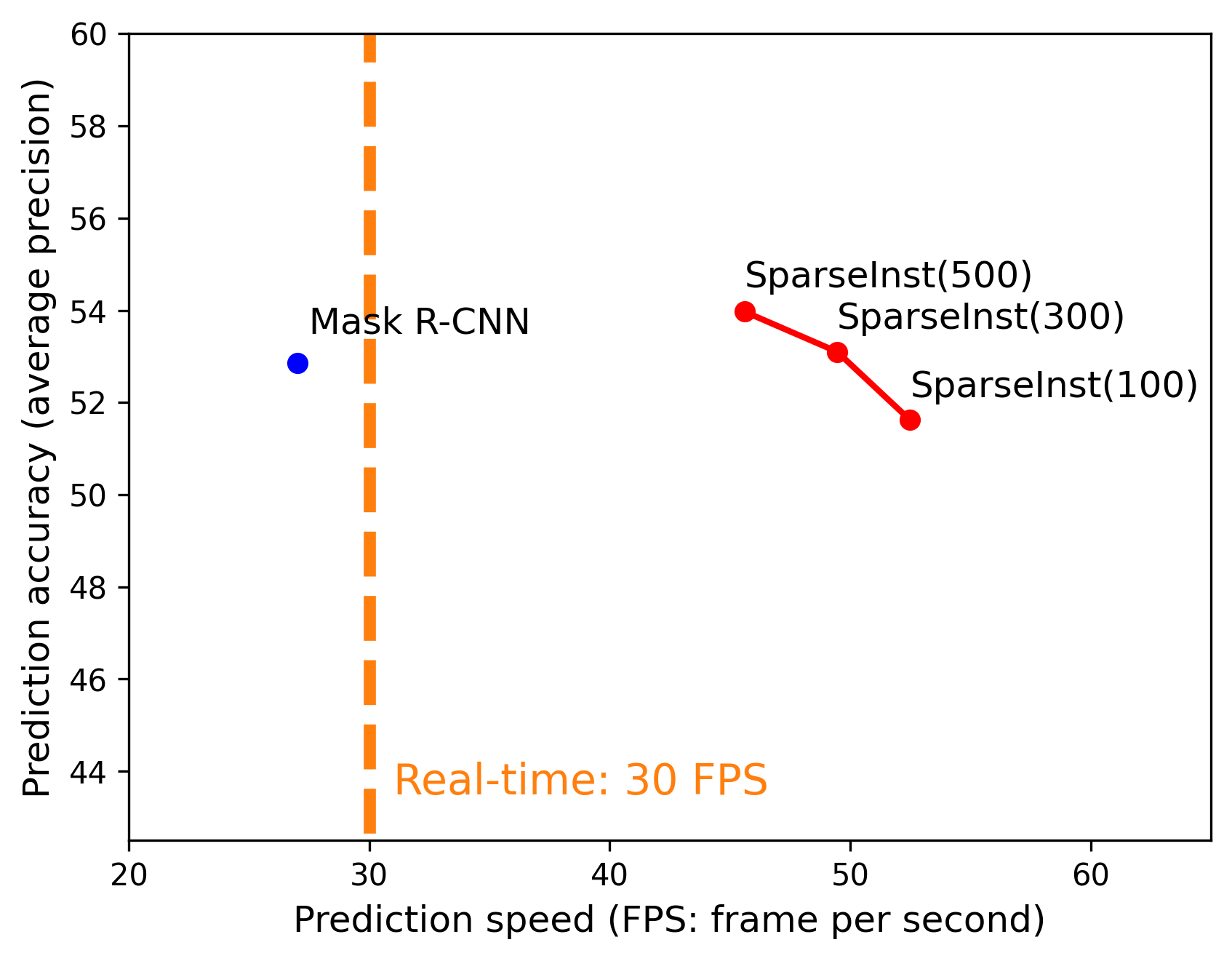}
  \caption{Speed and accuracy trade-off.}
  \Description{}
  \label{fig:speed_acc_tradeoff}
\end{figure}

Figure \ref{fig:speed_acc_tradeoff} shows the precision and speed trade-off of the SparseInst model and its comparison with Mask R-CNN. We used the default setting of Mask R-CNN to conduct training and testing as it achieves better performance than other experimental settings. Differently, SparseInst requires a predefined $N$ to determine the maximum number of masks and predictions per image. This hyperparameter not only affects the model's prediction accuracy but also its speed. A larger $N$ will slow down the process of bipartite matching during training and increase model complexity in the decoder part, therefore negatively affecting the model's efficiency during both training and testing. Here, we tested the model performance at different settings of $N$ (at 100, 300 and 500 respectively). It can be seen that as $N$ decreases, the model's prediction speed increases (x axis) but its predictive power (y axis) decreases (from 54\% at $N$=500 to 51\% at $N$=100). For Mask R-CNN, while its prediction accuracy is quite high, the speed is below the threshold of models that can be considered real-time. It is noteworthy that at both $N$=500 and $N$=300, SparseInst achieves better prediction accuracy than Mask R-CNN. This result verifies the importance of carefully setting values of hyperparameters according to data characteristics to achieve satisfying model performance.

\subsection{Prediction Results}
\begin{figure}
\centering
  \subcaptionbox{\label{fig:data_15_gt}}{\includegraphics[width=.48\linewidth]{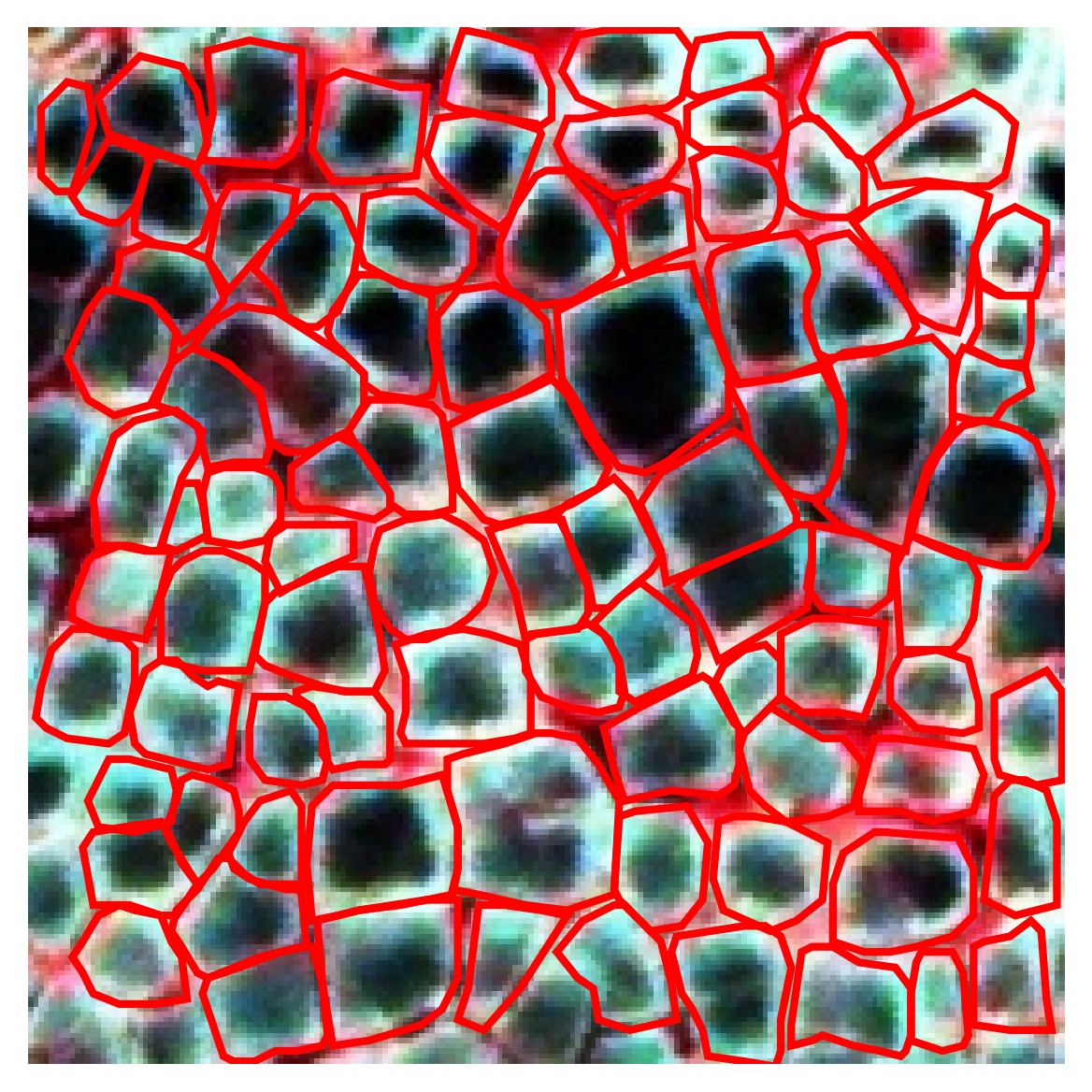}}
  \subcaptionbox{\label{fig:data_15_predict}}{\includegraphics[width=.48\linewidth]{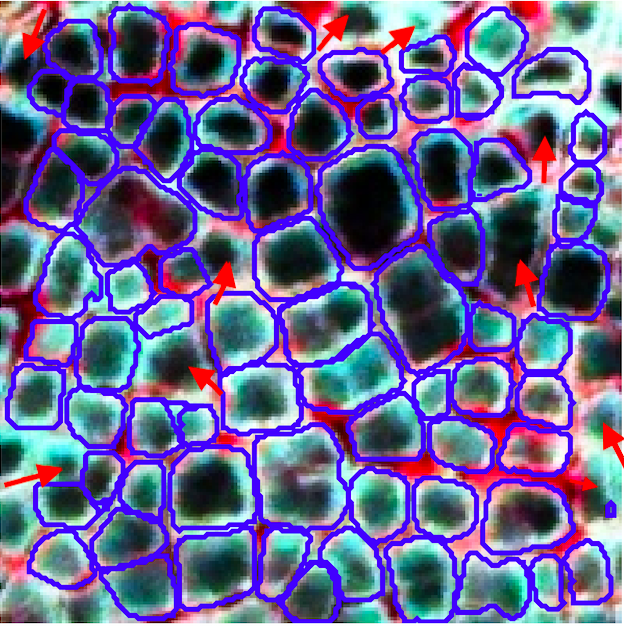}}
  \subcaptionbox{\label{fig:data_1131_gt}}{\includegraphics[width=.48\linewidth]{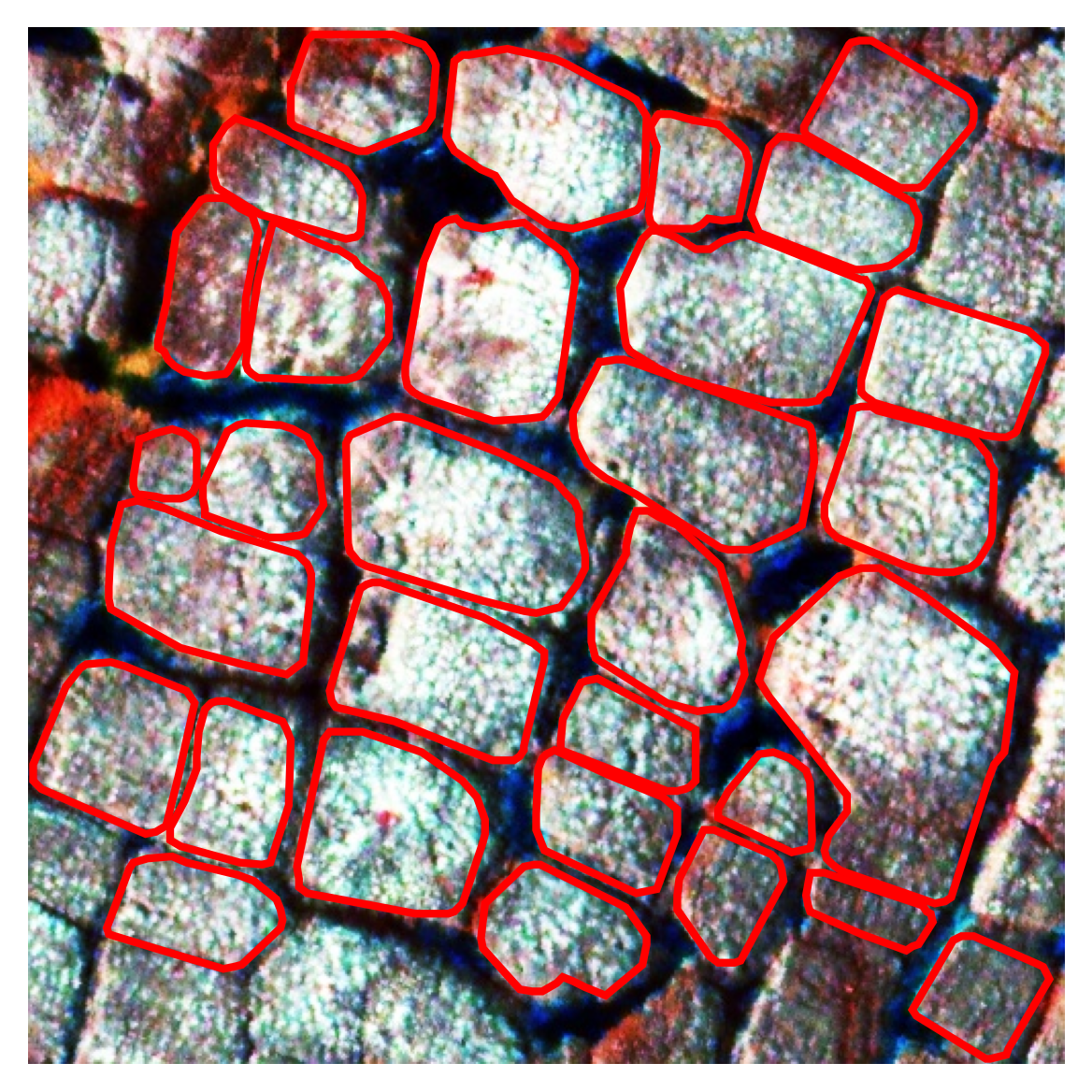}}
  \subcaptionbox{\label{fig:data_1131_predict}}{\includegraphics[width=.48\linewidth]{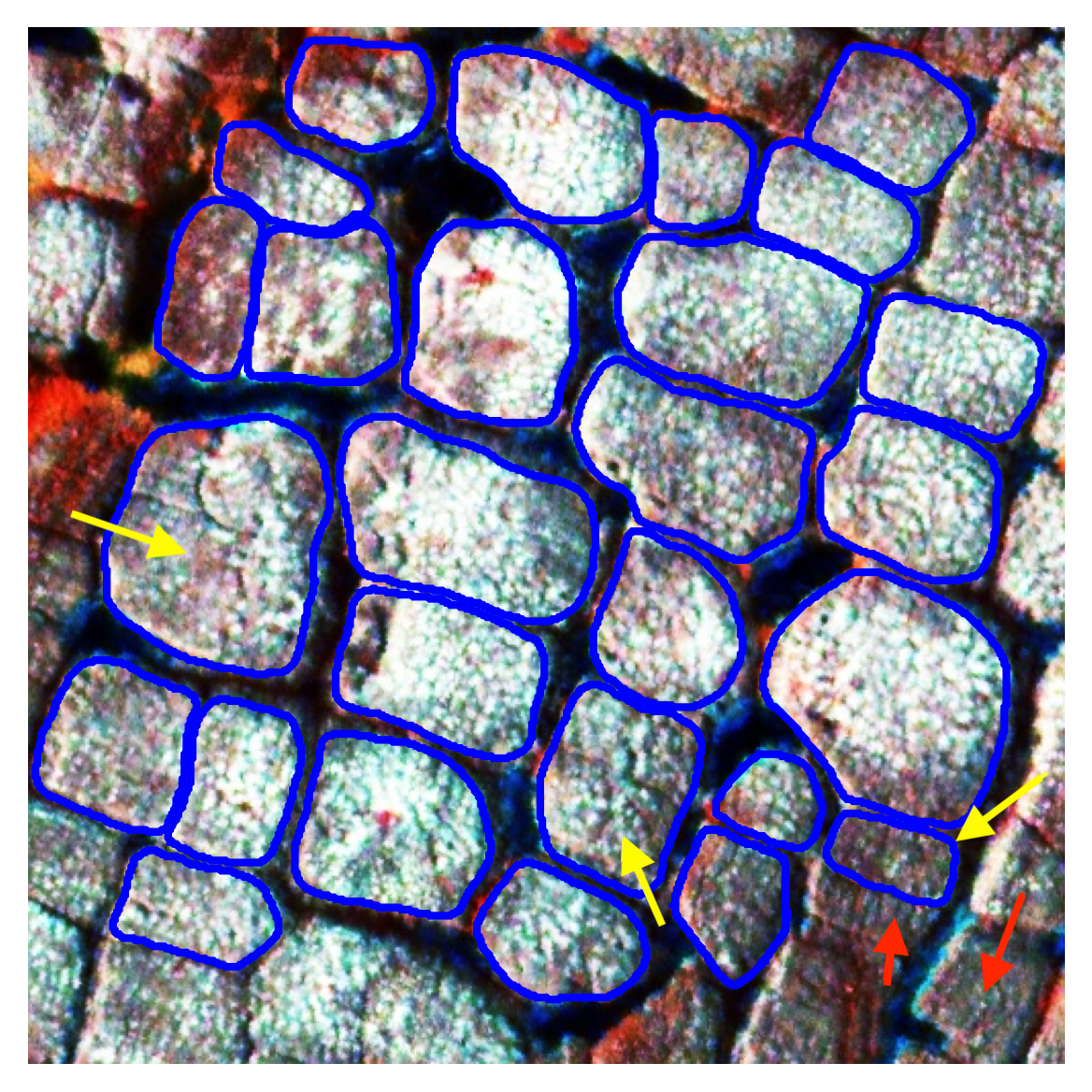}}
  \caption{Comparison between ground-truth (a and c) and model segmentation results (b and d). Red arrow: missing prediction; Yellow arrow: incorrect prediction}
  \Description{}
  \label{fig:data_sample}
\end{figure}

Figure \ref{fig:data_sample} illustrates segmentation results for two sample images. Figure \ref{fig:data_15_gt} and \ref{fig:data_1131_gt} provides the ground-truth labels of the IWPs. The ice-wedge polygons in these two images belong to two distinctive types of IWPs: low-centered (\ref{fig:data_15_gt}, \ref{fig:data_15_predict}) and high-centered (\ref{fig:data_1131_gt}, \ref{fig:data_1131_predict}). A preliminary analysis has also shown that when separating these feature types, thus making the segmentation task more challenging, the performance advantage of SparseInst over Mask R-CNN become even more dominant. This reflects the robustness of the SparseInst model in performing high-accuracy and real-time IWP segmentation.

Figure \ref{fig:data_15_predict} and \ref{fig:data_1131_predict} present the model prediction results for the two images to their left (\ref{fig:data_15_gt} and \ref{fig:data_1131_gt}). It can be seen that for smaller objects, although the predicted area is quite close to the ground-truth, the boundary line itself is not as smooth as the human labels, \ref{fig:data_15_predict}. This issue does not exist in segmentation results for large objects, \ref{fig:data_1131_predict}. The model did miss predictions for a few IWPs when there exist no clear boundaries around them (red arrows in \ref{fig:data_15_predict} and \ref{fig:data_1131_predict}). There are also incorrect predictions (yellow arrows in \ref{fig:data_1131_predict}); this is likely due to the semantically different concepts that expert annotators and the machine consider. Interestingly, the model can predict labels for some partial IWP near the border where is not labeled by experts. 

\section{Conclusion}
This paper introduces a real-time GeoAI workflow for segmenting an important permafrost feature, IWPs. Delineating their extent and qualifying their changes is critically important to understand Arctic warming and permafrost thaw and its impact to the Arctic environment, infrastructure, and people. Here, we adopt a light-weight instance segmentation model into the workflow and verify its good performance in terms of both prediction accuracy and speed. In the future, we will further improve both the training data to explicitly annotate multi-type IWPs, and also refine the model to improve its detection accuracy of small objects. 

\begin{acks}
This work is supported in part by the National Science Foundation under awards 2120943, 1853864, 1927872, and 2230034.
\end{acks}

\bibliographystyle{ACM-Reference-Format}
\bibliography{09-reference}


\begin{thebibliography}{23}

%%% the Web address.

\ifx \showCODEN    \undefined \def \showCODEN     #1{\unskip}     \fi
\ifx \showDOI      \undefined \def \showDOI       #1{#1}\fi
\ifx \showISBNx    \undefined \def \showISBNx     #1{\unskip}     \fi
\ifx \showISBNxiii \undefined \def \showISBNxiii  #1{\unskip}     \fi
\ifx \showISSN     \undefined \def \showISSN      #1{\unskip}     \fi
\ifx \showLCCN     \undefined \def \showLCCN      #1{\unskip}     \fi
\ifx \shownote     \undefined \def \shownote      #1{#1}          \fi
\ifx \showarticletitle \undefined \def \showarticletitle #1{#1}   \fi
\ifx \showURL      \undefined \def \showURL       {\relax}        \fi
% invisible to TeX
\providecommand\bibfield[2]{#2}
\providecommand\bibinfo[2]{#2}
\providecommand\natexlab[1]{#1}
\providecommand\showeprint[2][]{arXiv:#2}

\bibitem[Andersson et~al\mbox{.}(2020)]%
        {13_andersson2020deep}
\bibfield{author}{\bibinfo{person}{Tom Andersson}, \bibinfo{person}{Fruzsina
  Agocs}, \bibinfo{person}{Scott Hosking}, \bibinfo{person}{Mar{\'\i}a
  P{\'e}rez-Ortiz}, \bibinfo{person}{Brooks Paige}, \bibinfo{person}{Chris
  Russell}, \bibinfo{person}{Andrew Elliott}, \bibinfo{person}{Stephen Law},
  \bibinfo{person}{Jeremy Wilkinson}, \bibinfo{person}{Yevgeny Askenov},
  {et~al\mbox{.}}} \bibinfo{year}{2020}\natexlab{}.
\newblock \showarticletitle{Deep learning for monthly Arctic sea ice
  concentration prediction}. In \bibinfo{booktitle}{\emph{EGU General Assembly
  Conference Abstracts}}. \bibinfo{pages}{15481}.
\newblock


\bibitem[Bhuiyan et~al\mbox{.}(2020)]%
        {17_bhuiyan2020use}
\bibfield{author}{\bibinfo{person}{Md~Abul~Ehsan Bhuiyan},
  \bibinfo{person}{Chandi Witharana}, {and} \bibinfo{person}{Anna~K
  Liljedahl}.} \bibinfo{year}{2020}\natexlab{}.
\newblock \showarticletitle{Use of very high spatial resolution commercial
  satellite imagery and deep learning to automatically map ice-wedge polygons
  across tundra vegetation types}.
\newblock \bibinfo{journal}{\emph{Journal of Imaging}} \bibinfo{volume}{6},
  \bibinfo{number}{12} (\bibinfo{year}{2020}), \bibinfo{pages}{137}.
\newblock


\bibitem[Bianco et~al\mbox{.}(2018)]%
        {bianco2018benchmark}
\bibfield{author}{\bibinfo{person}{Simone Bianco}, \bibinfo{person}{Remi
  Cadene}, \bibinfo{person}{Luigi Celona}, {and} \bibinfo{person}{Paolo
  Napoletano}.} \bibinfo{year}{2018}\natexlab{}.
\newblock \showarticletitle{Benchmark Analysis of Representative Deep Neural
  Network Architectures}.
\newblock \bibinfo{journal}{\emph{IEEE access}}  \bibinfo{volume}{6}
  (\bibinfo{year}{2018}), \bibinfo{pages}{64270--64277}.
\newblock
\showISBNx{2169-3536}


\bibitem[Bolya et~al\mbox{.}(2019)]%
        {bolya2019yolacta}
\bibfield{author}{\bibinfo{person}{Daniel Bolya}, \bibinfo{person}{Chong Zhou},
  \bibinfo{person}{Fanyi Xiao}, {and} \bibinfo{person}{Yong~Jae Lee}.}
  \bibinfo{year}{2019}\natexlab{}.
\newblock \showarticletitle{Yolact: {{Real-time}} Instance Segmentation}. In
  \bibinfo{booktitle}{\emph{Proceedings of the {{IEEE}}/{{CVF}} International
  Conference on Computer Vision}}. \bibinfo{pages}{9157--9166}.
\newblock


\bibitem[Cheng et~al\mbox{.}(2022)]%
        {cheng2022sparse}
\bibfield{author}{\bibinfo{person}{Tianheng Cheng}, \bibinfo{person}{Xinggang
  Wang}, \bibinfo{person}{Shaoyu Chen}, \bibinfo{person}{Wenqiang Zhang},
  \bibinfo{person}{Qian Zhang}, \bibinfo{person}{Chang Huang},
  \bibinfo{person}{Zhaoxiang Zhang}, {and} \bibinfo{person}{Wenyu Liu}.}
  \bibinfo{year}{2022}\natexlab{}.
\newblock \showarticletitle{Sparse {{Instance Activation}} for {{Real-Time
  Instance Segmentation}}}. In \bibinfo{booktitle}{\emph{Proceedings of the
  {{IEEE}}/{{CVF Conference}} on {{Computer Vision}} and {{Pattern
  Recognition}}}}. \bibinfo{pages}{4433--4442}.
\newblock


\bibitem[Grosse et~al\mbox{.}(2016)]%
        {8_grosse2016changing}
\bibfield{author}{\bibinfo{person}{Guido Grosse}, \bibinfo{person}{Scott
  Goetz}, \bibinfo{person}{A~Dave McGuire}, \bibinfo{person}{Vladimir~E
  Romanovsky}, {and} \bibinfo{person}{Edward~AG Schuur}.}
  \bibinfo{year}{2016}\natexlab{}.
\newblock \showarticletitle{Changing permafrost in a warming world and
  feedbacks to the Earth system}.
\newblock \bibinfo{journal}{\emph{Environmental Research Letters}}
  \bibinfo{volume}{11}, \bibinfo{number}{4} (\bibinfo{year}{2016}),
  \bibinfo{pages}{040201}.
\newblock


\bibitem[He et~al\mbox{.}(2017)]%
        {maskrcnnhe2017mask}
\bibfield{author}{\bibinfo{person}{Kaiming He}, \bibinfo{person}{Georgia
  Gkioxari}, \bibinfo{person}{Piotr Doll{\'a}r}, {and} \bibinfo{person}{Ross
  Girshick}.} \bibinfo{year}{2017}\natexlab{}.
\newblock \showarticletitle{Mask r-cnn}. In
  \bibinfo{booktitle}{\emph{Proceedings of the IEEE international conference on
  computer vision}}. \bibinfo{pages}{2961--2969}.
\newblock


\bibitem[He et~al\mbox{.}(2016)]%
        {he2016deep}
\bibfield{author}{\bibinfo{person}{Kaiming He}, \bibinfo{person}{Xiangyu
  Zhang}, \bibinfo{person}{Shaoqing Ren}, {and} \bibinfo{person}{Jian Sun}.}
  \bibinfo{year}{2016}\natexlab{}.
\newblock \showarticletitle{Deep Residual Learning for Image Recognition}. In
  \bibinfo{booktitle}{\emph{Proceedings of the {{IEEE}} Conference on Computer
  Vision and Pattern Recognition}}. \bibinfo{pages}{770--778}.
\newblock


\bibitem[Hjort et~al\mbox{.}(2018)]%
        {4_hjort2018degrading}
\bibfield{author}{\bibinfo{person}{Jan Hjort}, \bibinfo{person}{Olli
  Karjalainen}, \bibinfo{person}{Juha Aalto}, \bibinfo{person}{Sebastian
  Westermann}, \bibinfo{person}{Vladimir~E Romanovsky},
  \bibinfo{person}{Frederick~E Nelson}, \bibinfo{person}{Bernd
  Etzelm{\"u}ller}, {and} \bibinfo{person}{Miska Luoto}.}
  \bibinfo{year}{2018}\natexlab{}.
\newblock \showarticletitle{Degrading permafrost puts Arctic infrastructure at
  risk by mid-century}.
\newblock \bibinfo{journal}{\emph{Nature communications}} \bibinfo{volume}{9},
  \bibinfo{number}{1} (\bibinfo{year}{2018}), \bibinfo{pages}{1--9}.
\newblock


\bibitem[Huang et~al\mbox{.}(2022)]%
        {lingcao2022accuracy}
\bibfield{author}{\bibinfo{person}{Lingcao Huang}, \bibinfo{person}{Trevor~C
  Lantz}, \bibinfo{person}{Robert~H Fraser}, \bibinfo{person}{Kristy~F Tiampo},
  \bibinfo{person}{Michael~J Willis}, {and} \bibinfo{person}{Kevin Schaefer}.}
  \bibinfo{year}{2022}\natexlab{}.
\newblock \showarticletitle{Accuracy, Efficiency, and Transferability of a Deep
  Learning Model for Mapping Retrogressive Thaw Slumps across the Canadian
  Arctic}.
\newblock \bibinfo{journal}{\emph{Remote Sensing}} \bibinfo{volume}{14},
  \bibinfo{number}{12} (\bibinfo{year}{2022}), \bibinfo{pages}{2747}.
\newblock


\bibitem[Li(2020)]%
        {li2020geoai}
\bibfield{author}{\bibinfo{person}{Wenwen Li}.}
  \bibinfo{year}{2020}\natexlab{}.
\newblock \showarticletitle{GeoAI: Where machine learning and big data converge
  in GIScience}.
\newblock \bibinfo{journal}{\emph{Journal of Spatial Information Science}}
  \bibinfo{number}{20} (\bibinfo{year}{2020}), \bibinfo{pages}{71--77}.
\newblock


\bibitem[Li and Hsu(2022)]%
        {li2022geoai}
\bibfield{author}{\bibinfo{person}{Wenwen Li} {and} \bibinfo{person}{Chia-Yu
  Hsu}.} \bibinfo{year}{2022}\natexlab{}.
\newblock \showarticletitle{GeoAI for Large-Scale Image Analysis and Machine
  Vision: Recent Progress of Artificial Intelligence in Geography}.
\newblock \bibinfo{journal}{\emph{ISPRS International Journal of
  Geo-Information}} \bibinfo{volume}{11}, \bibinfo{number}{7}
  (\bibinfo{year}{2022}), \bibinfo{pages}{385}.
\newblock


\bibitem[Li et~al\mbox{.}(2019)]%
        {li2019scaleaware}
\bibfield{author}{\bibinfo{person}{Yanghao Li}, \bibinfo{person}{Yuntao Chen},
  \bibinfo{person}{Naiyan Wang}, {and} \bibinfo{person}{Zhao-Xiang Zhang}.}
  \bibinfo{year}{2019}\natexlab{}.
\newblock \showarticletitle{Scale-{{Aware Trident Networks}} for {{Object
  Detection}}}. In \bibinfo{booktitle}{\emph{2019 {{IEEE}}/{{CVF International
  Conference}} on {{Computer Vision}} ({{ICCV}})}}.
  \bibinfo{publisher}{{IEEE}}, \bibinfo{address}{{Seoul, Korea (South)}},
  \bibinfo{pages}{6053--6062}.
\newblock
\showISBNx{978-1-72814-803-8}
\urldef\tempurl%
\url{https://doi.org/10.1109/ICCV.2019.00615}
\showDOI{\tempurl}


\bibitem[Lin et~al\mbox{.}(2017)]%
        {lin2017feature}
\bibfield{author}{\bibinfo{person}{Tsung-Yi Lin}, \bibinfo{person}{Piotr
  Doll{\'a}r}, \bibinfo{person}{Ross Girshick}, \bibinfo{person}{Kaiming He},
  \bibinfo{person}{Bharath Hariharan}, {and} \bibinfo{person}{Serge Belongie}.}
  \bibinfo{year}{2017}\natexlab{}.
\newblock \showarticletitle{Feature Pyramid Networks for Object Detection}. In
  \bibinfo{booktitle}{\emph{Proceedings of the {{IEEE}} Conference on Computer
  Vision and Pattern Recognition}}. \bibinfo{pages}{2117--2125}.
\newblock


\bibitem[NOAA(2019)]%
        {14_NOAA}
\bibfield{author}{\bibinfo{person}{NOAA}.} \bibinfo{year}{2019}\natexlab{}.
\newblock \showarticletitle{Developing artificial intelligence to find ice
  seals and polar bears from the sky}.
\newblock
  \bibinfo{journal}{\emph{https://www.fisheries.noaa.gov/feature-story/developing-artificial-intelligence-find-ice-seals-and-polar-bears-sky}}
  (\bibinfo{year}{2019}).
\newblock
\showISBNx{(Last access August.15, 2022)}


\bibitem[Perreault et~al\mbox{.}(2021)]%
        {perreault2021centerpoly}
\bibfield{author}{\bibinfo{person}{Hughes Perreault},
  \bibinfo{person}{Guillaume-Alexandre Bilodeau}, \bibinfo{person}{Nicolas
  Saunier}, {and} \bibinfo{person}{Maguelonne H{\'e}ritier}.}
  \bibinfo{year}{2021}\natexlab{}.
\newblock \showarticletitle{{{CenterPoly}}: Real-Time Instance Segmentation
  Using Bounding Polygons}.
\newblock \bibinfo{journal}{\emph{arXiv:2108.08923 [cs]}}
  (\bibinfo{date}{Sept.} \bibinfo{year}{2021}).
\newblock
\showeprint[arxiv]{2108.08923}~[cs]


\bibitem[Schuur et~al\mbox{.}(2015)]%
        {schuur2015climate}
\bibfield{author}{\bibinfo{person}{Edward~AG Schuur}, \bibinfo{person}{A~David
  McGuire}, \bibinfo{person}{Christina Sch{\"a}del}, \bibinfo{person}{Guido
  Grosse}, \bibinfo{person}{Jennifer~W Harden}, \bibinfo{person}{Daniel~J
  Hayes}, \bibinfo{person}{Gustaf Hugelius}, \bibinfo{person}{Charles~D Koven},
  \bibinfo{person}{Peter Kuhry}, \bibinfo{person}{David~M Lawrence},
  {et~al\mbox{.}}} \bibinfo{year}{2015}\natexlab{}.
\newblock \showarticletitle{Climate change and the permafrost carbon feedback}.
\newblock \bibinfo{journal}{\emph{Nature}} \bibinfo{volume}{520},
  \bibinfo{number}{7546} (\bibinfo{year}{2015}), \bibinfo{pages}{171--179}.
\newblock


\bibitem[Song et~al\mbox{.}(2019)]%
        {15_song2019patch}
\bibfield{author}{\bibinfo{person}{Hunsoo Song}, \bibinfo{person}{Yonghyun
  Kim}, {and} \bibinfo{person}{Yongil Kim}.} \bibinfo{year}{2019}\natexlab{}.
\newblock \showarticletitle{A patch-based light convolutional neural network
  for land-cover mapping using Landsat-8 images}.
\newblock \bibinfo{journal}{\emph{Remote Sensing}} \bibinfo{volume}{11},
  \bibinfo{number}{2} (\bibinfo{year}{2019}), \bibinfo{pages}{114}.
\newblock


\bibitem[Udawalpola et~al\mbox{.}(2022)]%
        {udawalpola2022optimal}
\bibfield{author}{\bibinfo{person}{Mahendra~R Udawalpola},
  \bibinfo{person}{Amit Hasan}, \bibinfo{person}{Anna Liljedahl},
  \bibinfo{person}{Aiman Soliman}, \bibinfo{person}{Jeffrey Terstriep}, {and}
  \bibinfo{person}{Chandi Witharana}.} \bibinfo{year}{2022}\natexlab{}.
\newblock \showarticletitle{An Optimal GeoAI Workflow for Pan-Arctic Permafrost
  Feature Detection from High-Resolution Satellite Imagery}.
\newblock \bibinfo{journal}{\emph{Photogrammetric Engineering \& Remote
  Sensing}} \bibinfo{volume}{88}, \bibinfo{number}{3} (\bibinfo{year}{2022}),
  \bibinfo{pages}{181--188}.
\newblock


\bibitem[Wang et~al\mbox{.}(2020)]%
        {NEURIPS2020_cd3afef9}
\bibfield{author}{\bibinfo{person}{Xinlong Wang}, \bibinfo{person}{Rufeng
  Zhang}, \bibinfo{person}{Tao Kong}, \bibinfo{person}{Lei Li}, {and}
  \bibinfo{person}{Chunhua Shen}.} \bibinfo{year}{2020}\natexlab{}.
\newblock \showarticletitle{{{SOLOv2}}: {{Dynamic}} and Fast Instance
  Segmentation}. In \bibinfo{booktitle}{\emph{Advances in Neural Information
  Processing Systems}}, \bibfield{editor}{\bibinfo{person}{H.~Larochelle},
  \bibinfo{person}{M.~Ranzato}, \bibinfo{person}{R.~Hadsell},
  \bibinfo{person}{M.~F. Balcan}, {and} \bibinfo{person}{H.~Lin}} (Eds.),
  Vol.~\bibinfo{volume}{33}. \bibinfo{publisher}{{Curran Associates, Inc.}},
  \bibinfo{pages}{17721--17732}.
\newblock


\bibitem[Wu et~al\mbox{.}(2019)]%
        {wu2019detectron2}
\bibfield{author}{\bibinfo{person}{Yuxin Wu}, \bibinfo{person}{Alexander
  Kirillov}, \bibinfo{person}{Francisco Massa}, \bibinfo{person}{Wan-Yen Lo},
  {and} \bibinfo{person}{Ross Girshick}.} \bibinfo{year}{2019}\natexlab{}.
\newblock \bibinfo{title}{Detectron2}.
\newblock
  \bibinfo{howpublished}{\url{https://github.com/facebookresearch/detectron2}}.
\newblock


\bibitem[Zhang and Zhang(2009)]%
        {Zhang2009}
\bibfield{author}{\bibinfo{person}{Ethan Zhang} {and} \bibinfo{person}{Yi
  Zhang}.} \bibinfo{year}{2009}\natexlab{}.
\newblock \bibinfo{booktitle}{\emph{Average Precision}}.
\newblock \bibinfo{publisher}{Springer US}, \bibinfo{address}{Boston, MA},
  \bibinfo{pages}{192--193}.
\newblock
\showISBNx{978-0-387-39940-9}
\urldef\tempurl%
\url{https://doi.org/10.1007/978-0-387-39940-9_482}
\showDOI{\tempurl}


\bibitem[Zhao et~al\mbox{.}(2017)]%
        {zhao2017pyramida}
\bibfield{author}{\bibinfo{person}{Hengshuang Zhao}, \bibinfo{person}{Jianping
  Shi}, \bibinfo{person}{Xiaojuan Qi}, \bibinfo{person}{Xiaogang Wang}, {and}
  \bibinfo{person}{Jiaya Jia}.} \bibinfo{year}{2017}\natexlab{}.
\newblock \showarticletitle{Pyramid Scene Parsing Network}. In
  \bibinfo{booktitle}{\emph{Proceedings of the {{IEEE}} Conference on Computer
  Vision and Pattern Recognition}}. \bibinfo{pages}{2881--2890}.
\newblock


\end{thebibliography}

\end{document}